\documentclass{article}
\usepackage{spconf,amsmath,epsfig}
\usepackage{adjustbox, caption}
\usepackage{amssymb}
\usepackage{pifont}
\usepackage[dvipsnames]{xcolor}
\usepackage{booktabs}
\usepackage{multirow}
\usepackage{url}
\usepackage[compress]{cite}

\newcommand{\cmark}{\textcolor{Green}{\text{\ding{51}}}}%
\newcommand{\xmark}{\textcolor{Red}{\text{\ding{55}}}}%

\let\OLDthebibliography\thebibliography
\renewcommand\thebibliography[1]{
  \OLDthebibliography{#1}
  \setlength{\parskip}{0pt}
  \setlength{\itemsep}{0pt plus 0.3ex}
}

\begin{document}\sloppy

\def\x{{\mathbf x}}
\def\L{{\cal L}}

\title{\textit{Protégé}: Learn and Generate Basic Makeup Styles with Generative Adversarial Networks (GANs)}
%
\name{Jia Wei Sii \quad Chee Seng Chan}
\address{Center of Signal and Image Processing (CISiP), Universiti Malaya, 50603 Kuala Lumpur, Malaysia}

\twocolumn[{%
\renewcommand\twocolumn[1][]{#1}%
\maketitle
}]


%
\begin{abstract}
Makeup is no longer confined to physical application; people now use mobile apps to digitally apply makeup to their photos, which they then share on social media. However, while this shift has made makeup more accessible, designing diverse makeup styles tailored to individual faces remains a challenge. This challenge currently must still be done manually by humans. Existing systems, such as makeup recommendation engines and makeup transfer techniques, offer limitations in creating innovative makeups for different individuals ``intuitively''---significant user effort and knowledge needed and limited makeup options available in app. Our motivation is to address this challenge by proposing \textbf{\textit{Protégé}}, a new makeup application, leveraging recent generative model---GANs to learn and automatically generate makeup styles. This is a task that existing makeup applications (i.e., makeup recommendation systems using expert system and makeup transfer methods) are unable to perform. Extensive experiments has been conducted to demonstrate the capability of \textit{Protégé} in learning and creating diverse makeups, providing a convenient and intuitive way, marking a significant leap in digital makeup technology! 
\end{abstract}



\section{Introduction}
\label{introduction}
Existing digital makeup applications have shown their capabilities in photo editing and beauty applications. They often fall into three categories which are (1) manual makeup applications such as  Meitu \cite{meitu}, ModiFace \cite{modiface2024}, Perfect365 \cite{perfect3652024}, BeautyPlus \cite{beautyplus2024}, YouCamMakeup \cite{youcam2024}, and FaceTune2 \cite{facetune2024}, (2) makeup recommendation systems using expert system \cite{tong2007example,guo2009digital,liu2016makeup,gulati2023beautifai,yan2023beautyrec} and (3) makeup transfer methods \cite{alashkar2017examples,scherbaum2011computer,liu2014wow,alashkar2017rule,nguyen2017smart}. 

These three categories offer makeup functionalities; however, they share the common drawbacks---they fail to replicate makeup applications functionalities and characteristics that a human makeup artist would have. Specifically, a human makeup artist does innovative and tailored makeups for different individuals in an intuitive manner. “Intuitive” in this context means a makeup artist would decide and create makeup based on their makeup experiences and aesthetic instincts. Existing technologies lack these intuition and innovation capabilities that are necessary to meet the increasingly sophisticated demands of today’s consumers. 

This shortfall is particularly evident in conventional digital makeup systems, which include manual makeup enhancement applications, makeup recommendation systems, and makeup transfer systems. Manual makeup applications such as MeiTu, FaceTune, YouCam Perfect offer users an array of makeup styles to experiment with. Yet, this requires significant user effort and knowledge as it involves trial and error in mixing and matching makeup for different facial features, which is not intuitive. Moreover, this approach is often hampered by the limited makeup options available within the apps, restricting the innovation in makeup applications. Consequently, users frequently settle for less than satisfactory makeup choices.

\begin{figure}[ht]
	\centering 
	\includegraphics[width=0.48\textwidth]{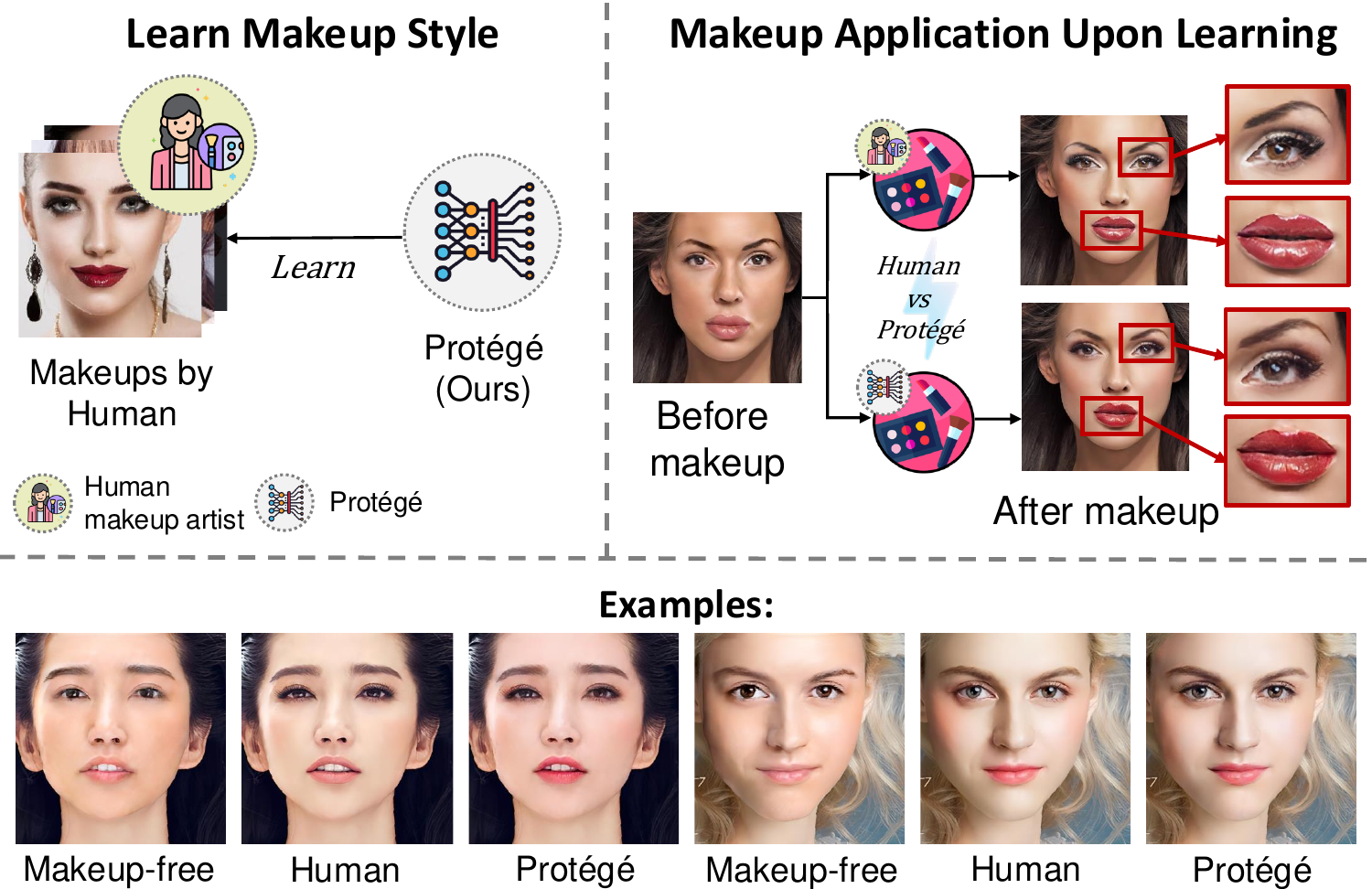}
	\caption{\textbf{Overview of \textit{Protégé}.} (Top left) \textit{Protégé} learns basic makeup styles from a curated makeup dataset. (Top right) After training, the model applies its learned styles and knowledge to generate and apply makeup on individual faces. (Bottom) Additional examples of makeup generated by \textit{Protégé}.}
	\label{fig:overview}
\end{figure}

Makeup recommendation systems with expert system aim to simplify user experience by automating makeup selections through rule-based algorithms. Makeup recommendation systems aim to streamline the user experience by automating makeup selections through rule-based algorithms. These algorithms operate on predefined, expert-derived rules—for example, recommending peach and coral shades for users with a warm undertone or suggesting specific eyeliner or lipstick options based on facial features. Basically, to make this makeup recommendation system effective, these rules must be made of all possible kinds of cosmetic options in the world. While this automation simplifies the procedures for the end users, it transfers the burden to industry experts who must develop extensive, detailed rules for makeup combinations, effectively replacing rather than reducing the effort required, not even to say the capability to build this extensive expert system is impractical. Moreover, these systems lack the intuitive judgment that characterizes professional artistry, where decisions are based on experience and instinct rather than rigid rules.

Makeup transfer, on the other hand, was designed to allow users to preview makeup on their faces digitally before actual application. These systems are often integrated in the existing mobile makeup applications to allow the users to direct transfer the existing makeup onto their faces. However, as we discussed, this method only provides limited choices of makeup options for the users to choose. Besides, this process merely replicates existing makeup from a reference image, without creating adaptive and tailored makeups automatically, losing the intuition and innovation that a professional artist would offer. 

Responding to these challenges, we introduce \textbf{\textit{Protégé}}, a groundbreaking idea, the first kind of basic makeup application. Unlike existing systems, \textbf{\textit{Protégé}} learns and generates distinct makeup for individual intuitively.

As depicted in Fig. \ref{fig:overview}, the context ‘Learn’ refers to training our makeup model- \textit{Protégé}, to be a makeup artist that can do basic makeup like human. Following our designed methodology, \textit{Protégé} leans basic makeup styles from a basic makeup dataset. This training mirrors the training phase of a human makeup artist where each makeup artist would undergo at the beginning of their career. Upon training, the model is expected to equip with the makeup styles and knowledge it obtained from the dataset and the training phase. Whereas for the context ‘Generate’, it means that our model, \textit{Protégé} applies makeup to the face of the individual based on its learning. This makeup is generated intuitively, akin to the approach of a human artist, also, it is distinct with different face inputted. 

To facilitate this learning and generation process, \textit{Protégé} is equipped by our designed makeup inpainting algorithm that has repurposed from existing image inpainting techniques. Makeup inpainting algorithm allows \textit{Protégé} to learn the overall makeup style and makeup knowledge from a makeup dataset where we desire it to absorb. During the makeup application, this algorithm leads \textit{Protégé} to utilize information from the input face and surrounding information, then fill a makeup on face, completing a makeup generation process. This method opens a new avenue for the community by not only going beyond those makeup applications that offer only limited choices of makeup on digital platform but also proposing a human-like makeup application technique. Besides, it also opens the possibility to allow the makeup industries to train their customized digital makeup artist with their specific cosmetic styles. Our key contributions are: 
\begin{enumerate}
    \item We propose \textbf{\textit{Protégé}}, a basic makeup generation model that mimics human makeup artist to intuitively generate distinct and innovative makeup for different individuals.
    \item We introduce a GAN-based \textit{makeup inpainting} algorithm, learning and applying makeup that complements the input face while preserving the original face identity.
\end{enumerate}

\begin{table*}[t]
\centering
\adjustbox{max width=\textwidth}{
    \begin{tabular}{l c c c c c} 
     \hline
      & Manual Selection & Rule-based & Makeup Transfer & \textbf{\textit{Protégé}} \\
     \hline
     Require Makeup Dataset for Training & \cmark & \cmark & \cmark & \cmark\\
     Independent on User's Knowledge & \xmark & \cmark & \cmark & \cmark\\
     Independent on Expert Knowledge & \cmark & \xmark & \cmark & \cmark\\
     Independent on Rule Quality & \cmark & \xmark & \cmark & \cmark\\
     Independent on Reference Image & \xmark & \cmark & \xmark & \cmark\\
     Easy Maintenance and Trend Updates & \cmark & \xmark & \cmark & \cmark\\
     Flexible \& Intuitive Makeup Generation & \xmark & \xmark & \xmark & \cmark \\
     \hline
    \end{tabular}
    }
\caption{\textbf{Comparison of our method, \textit{Protégé} with existing methods.} Our method simplifies the inputs for the makeup processes, while generating intuitive makeup within a desired style, where the makeup is evaluated in our Experiments section.}
\label{tab:related_work}
\end{table*}

\section{Related Work}
\label{related_work}
\textbf{Makeup Applications.}
As summarized in Tab. \ref{tab:related_work}, the field of digital makeup application has evolved through several stages, beginning with simple filter-based manual makeup enhancement applications \cite{perera2021virtual,meitu,beautyplus2024,modiface2024,perfect3652024,youcam2024,facetune2024} that rely heavily on user input, advancing through rule-based systems \cite{alashkar2017examples,liu2014wow,alashkar2017rule,nguyen2017smart} that automate some decisions but lack flexibility, to more sophisticated makeup transfer techniques \cite{tong2007example,guo2009digital,scherbaum2011computer,liu2016makeup,gulati2023beautifai,yan2023beautyrec}. These developments, while each marking an improvement over previous methods, often do not fully address the need for intuitive and innovative makeup application, causing a limited range and diversity of makeups available to end users. An intuitive and innovative makeup application reduces human efforts in digital makeup application and provides a useful tool in creating tailored makeup that is aligned with our specified makeup style. 

\textbf{Image Inpainting.}
Image inpainting \cite{suvorov2022resolution,li2022mat,li2022misf,liu2022reduceput,zheng2022bridgingtfill,lugmayr2022repaint,fcf} is a technique originally developed to restore damaged parts of images or fill in missing areas within digital photographs. The primary objective of traditional image inpainting is to reconstruct coherent image regions with plausible textures and details that are visually seamless and consistent with the surrounding areas. We appreciate the concept of image inpainting, implementing the image inpainting techniques to integrate in facial makeup applications. We aim to repurpose this technique to become a new makeup algorithm so it can mimic human makeup artist to be able to create innovative makeup intuitively for different individuals. 

\section{Methodology}
\begin{figure*}[ht]
	\centering 
	\includegraphics[width=1\textwidth]{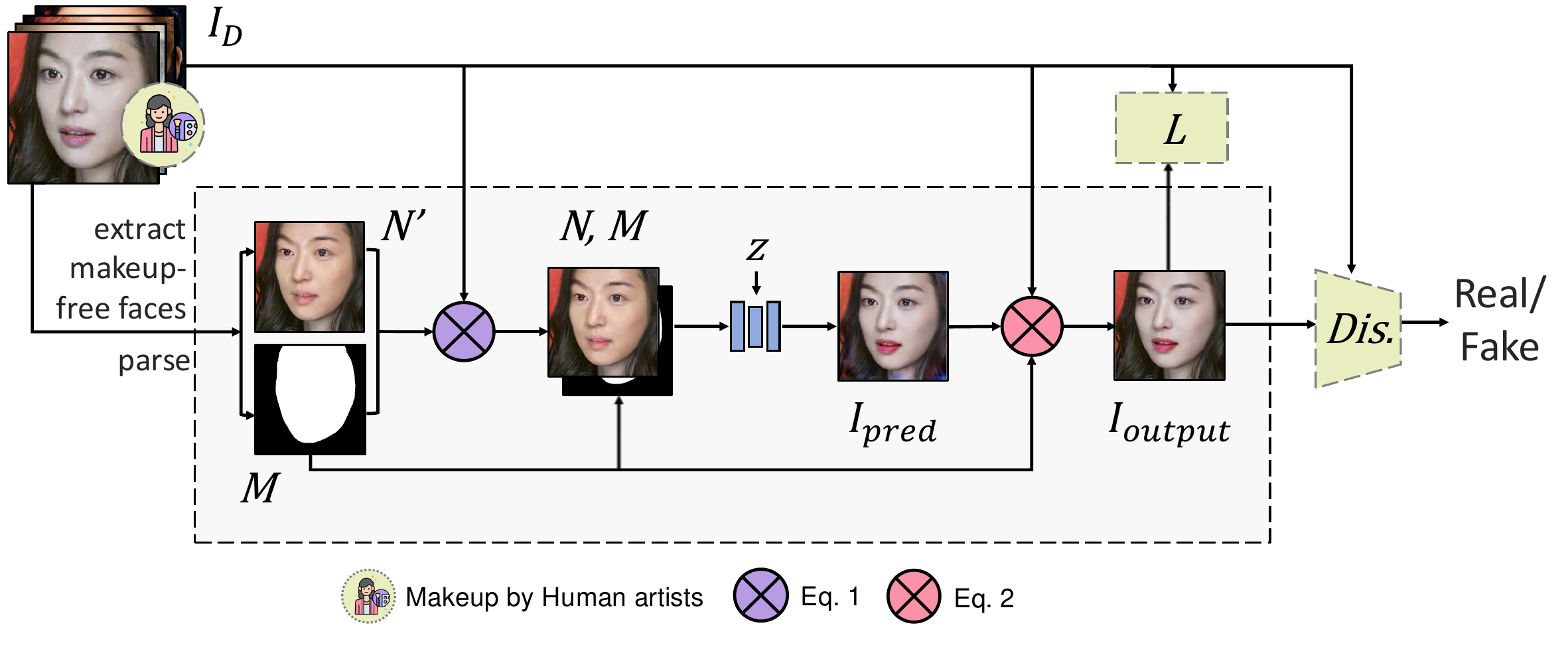}
	\caption{\textbf{Methodology Overview of \textit{Protégé} Makeup Learning and Generation Process.}
This figure illustrates the key stages of the process in \textit{Protégé}. First, a binary Region of Interest (ROI) mask is created to identify facial areas for makeup. \textit{Protégé} then comprehends the makeup style by generating a makeup-free face and blending it with the original image. It synthesizes a diverse range of makeup styles via a GAN-based generator, ensuring high-quality, personalized makeup while preserving facial identity. Finally, makeup is seamlessly integrated with the untouched non-facial areas to produce the final output image. Loss functions are employed to ensure stylistic accuracy and maintain the authenticity of the makeup. This framework enables \textit{Protégé} to effectively learn and generate innovative makeup styles tailored to individual faces.}

	\label{fig:makeup_framework}
\end{figure*}

To develop a sophisticated digital basic makeup application system---\textit{Protégé}, that autonomously learns from a vast dataset with basic makeups and then applies this knowledge to generate distinct, dataset-aligned makeups that are tailored to individual facial features. As illustrated in Fig. \ref{fig:makeup_framework}, this system integrates generative adversarial networks (GANs) \cite{goodfellow2014generative} and repurposes image inpainting method \cite{fcf} to achieve a level of intuitiveness and makeup innovation in digital makeup applications that mirrors human’s makeup generation capabilities. The reason for using Generative Adversarial Networks (GANs) instead of stable diffusion models in this work is that GANs focus solely on learning from the given dataset. In contrast, stable diffusion models typically rely on pre-trained models, which incorporate existing knowledge beyond the dataset. This external knowledge can affect the quality of the basic makeup generation we aim to achieve.

\subsection{Preliminary}
The foundation of \textit{Protégé}’s capabilities is its robust dataset of makeup images, designated as $I_{\text{D}}$. These images, all crafted by professional human makeup artists, represent a wide range of makeups with different artistic expressions, poses and lighting. This dataset is pivotal for our model to analyze and internalize the makeup distribution which represents the basic makeup style that we want our model to learn, allowing it to synthesize bespoke makeup styles that are both innovative and aligned with that particular makeup style which is basic makeup style in our case.

\subsection{Makeup learning and generation via makeup inpainting}
\textit{Protégé} commences its learning process by deeply analyzing a dataset of basic makeup images. This initial stage is crucial as it allows the system to absorb and understand our specified makeup style, makeup knowledge, and the specific interplay of colors and textures from the dataset. The goal here is to build a framework that captures the essence of the makeup style to facilitate the generation of innovative while style-aligned makeup applications on new subjects. Next, we elaborate on the makeup inpainting algorithm for makeup learning and generation.

\subsubsection{Identify the area for makeup}
To understand the location where the makeup should be applied, the first technical step in our makeup inpainting algorithm involves defining the regions of interest on the face where makeup will be applied. This is achieved by extracting a binary Region of Interest (ROI) mask $M$ using advanced face-parsing technique \cite{yu2018bisenet}. This mask helps in delineating facial features that are critical for personalized makeup application. In this mask, facial areas designated for makeup application are marked with a value of 1, indicating that they are active regions for makeup application. Non-facial areas and regions not targeted for makeup, such as the hair, neck, background, are marked with a value of 0, indicating that these areas should remain unaffected during the makeup application process. This binary distinction helps in focusing the application algorithms specifically where makeup is intended, ensuring clarity and precision in the final output.

\subsubsection{Comprehend makeup by applying makeup on makeup-free face and assessing the makeup}
The core of \textit{Protégé}'s functionality is its makeup comprehending process. This process begins with the extraction of a makeup-free version of the face from each image from the dataset, using a sophisticated cross-image style transfer technique, known as LADN\cite{gu2019ladn}. This technique strips the original makeup from the face, revealing its natural state and providing a clean slate for makeup application $N'=\mathrm{LADN}(I_{\text{D}})$. However, the resultant makeup-free image $N'$ is not only devoid of makeup on the face but also free of any makeup traces in non-facial areas such as the neck, hair, and background, which is undesired.

To preserve the integrity of non-facial areas as the original image in the dataset during the extraction of makeup-free face, $N'$ is blended with the corresponding image from the dataset, $I_{\text{D}}$  using the previously defined mask $M$. The resulting composite image $N$, which is our desired resultant makeup-free image, is then concatenated with $M$ to form a four-channel input $I$, which is utilized in the subsequent inpainting stages, where the equation as:
\begin{align} \label{eq:1}
    N &= N' \times M + I_{\text{D}} \times (1 - M), \\
    I &= \mathrm{concat}(N, M) \nonumber
\end{align}

With the preparatory stages complete, \textit{Protégé} proceeds to the actual application of makeup. Emulating the techniques of human makeup artists, our model, inspired by image inpainting technique, repurposes the image inpainting to intuitively apply makeup across the defined facial areas. In the following stages, \textit{Protégé} reflects the makeup procedure that a human would do on the targeted face. 

\textbf{(a) Process the prime information in \textit{N} for makeup.} This process reflects the human behavior of understanding targeted facial features before makeup application. Upon having the ingredient ready, the makeup comprehending process is continued with the critical encoding of $I$, which is the concatenated input of the makeup-free face $N$ and the ROI mask $M$. This initial encoding stage is crucial as it ensures that all the relevant information, especially the information from the masked region (i.e., face) is considered in the makeup application process.

Note: Conventional image inpainting techniques treat the masked region as a void, filling it in solely with information from the surrounding areas, as they do not expect any usable data from within the masked region itself. In contrast, our Makeup Inpainting algorithm considers the masked region as a crucial source of information for the makeup application process. It leverages both the surrounding details from the unmasked areas and the data within the masked region itself.

During the entire input processing, \textit{Protégé} includes facial identity. This approach mirrors the procedure before a makeup artist starts their makeup application-assessing a client’s facial features, choosing enhancements that suit the client.

\textbf{(b) Fine-grained and diverse makeup synthesis.} \textit{Protégé} uses a sophisticated mapping network to convert a noise latent vector into a detailed latent space. This space effectively captures the diversity of makeup styles contained within the training dataset. Subsequently, a StyleGAN2-based generator refines these inputs into a coherent makeup application in a coarse-to-fine manner. This step-by-step refinement process ensures that each layer of makeup is applied with increasing precision, reflecting the intricate details that characterize professional makeup applications. The results $I_{\text{pred}}$, is a high-quality representation of how the makeup looks on the user’s face, combining artistic flair with technical precision. 

Moreover, during the makeup synthesis, \textit{Protégé} maintains the facial identity, ensuring that any makeup enhancements complement rather than obscure the face of the subject. This approach mirrors the precision with which a makeup artist conducts makeup applications without altering the face identity. This capability is achieved using generative adversarial networks (GANs), which are designed to learn detailed and nuanced representations of facial features. GANs effectively capture and reinforce the unique characteristics of each face, allowing for personalized makeup applications that maintain the integrity of the original facial structure. By integrating this advanced technology, \textit{Protégé} ensures that the makeup application enhances the individual's appearance while staying true to their natural identity, much like the bespoke services offered by skilled human makeup artists. 

\textbf{(c) Preserve the integrity of non-facial areas after makeup application.} Note: This scenario would not occur in human makeup artist applications but is needed to be addressed in digital contexts.
After the initial application of makeup, it is imperative to integrate the newly applied cosmetics seamlessly with the non-facial areas of the original image $I_{\text{D}}$. This integration is meticulously achieved by using the ROI mask $M$ to blend the makeup-applied regions of $I_{\text{pred}}$ with the untouched areas of $I_{\text{D}}$. This careful blending ensures the non-targeted regions remain its integrity as the $I_{\text{D}}$. This final composite image $I_{\text{output}}$, is the product ready for professional evaluation and further refinement if necessary. The equation is written as below:  
\begin{equation} \label{eq:2}
    I_{\text{output}} = I_{\text{pred}} \times M + I_{\text{D}} \times (1 - M).
\end{equation}

\subsection{Makeup's checking and guarding.}
The final step in the \textit{Protégé}’s process involves rigorous quality assurance to ensure that the makeup applied not only meets the intended aesthetic standards but also remains true to the makeup style identified in the dataset. This is accomplished through a set of meticulously calibrated loss functions during the training phase. The non-saturating cross-entropy loss ensures that the generated makeup aligns closely with the makeup style in the dataset. The loss is implemented, where the input is either dataset's image $D$ or the final output $I_{\text{output}}$, based on the training phase. Whereas the high receptive perceptual loss LHRFPL fine-tunes the details for accuracy. It is achieved by calculating the $L_{\text{2}}$ distance between the final output  $I_{\text{output}}$ and $I_{\text{D}}$ in feature space, which is formulated as \(L_2 = \| I_{\text{output}} - I_D \|_2 \). Additionally, a pixel-wise reconstruction loss $L_{\text{rec}}$ assesses the overall quality, adjusting ensure that the final product is not only visually appealing but also authentic to the desired trend. It is achieved by calculating between $I_{\text{output}}$ and $I_{\text{D}}$, formulated as $L_{\text{rec}}$, which is formulated as \(L_rec = \| I_{\text{output}} - I_D \|_1 \).

The total loss function is formulated as:
$\sum L = L_{\text{adv}} + \lambda_{\text{HRFPL}} L_{\text{HRFPL}} + \lambda_{\text{rec}} L_{\text{rec}}$.

The generator and discriminator are trained adversarially, with empirically set loss weights \cite{fcf}: \(\lambda_{\text{rec}} = 10\) and \(\lambda_{\text{HRFPL}} = 5\).

\begin{figure*}[ht]
	\centering 
	\includegraphics[width=1\textwidth]{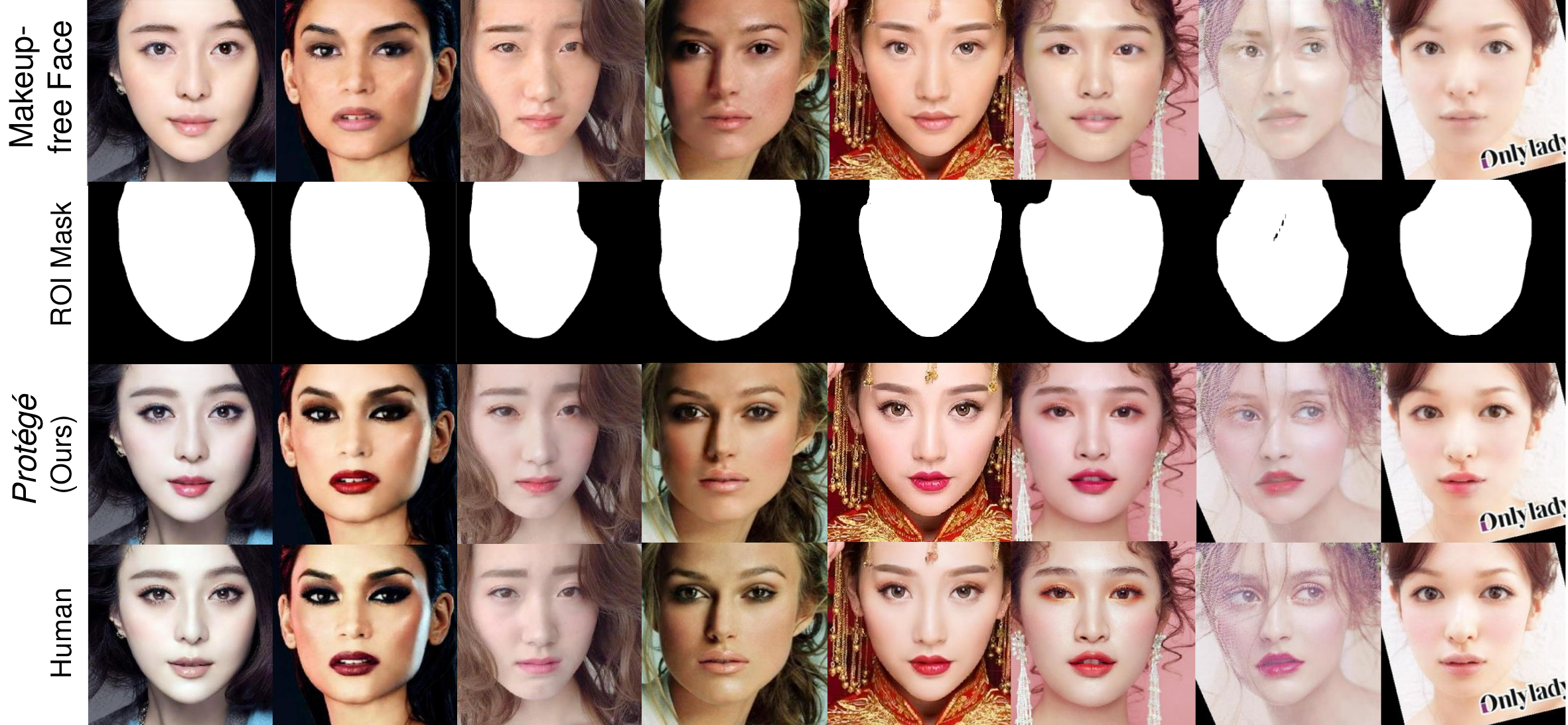}
	\caption{Visual Comparison between generated makeup by \textit{Protégé} and makeup done by human makeup artist.}
	\label{fig:qual_results}
\end{figure*}
\section{Experiments}

\textbf{Baselines.}
We evaluate our methodology from two perspectives: \textbf{(i) Makeup style learning:} To evaluate if the makeup applied on the targeted face is aligned with our desired style (i.e., the basic makeup style in dataset), we benchmark our set of resulting makeups with the dataset where it contains the basic style that we desire to follow. \textbf{(ii) Face Identity Preservation:} To evaluate if the face identity of the target subject is preserved upon makeup application by our \textit{Protégé}, we evaluate the resulting face after makeup applied with the makeup-free face before makeup. This evaluation is critical because we are repurposing from typical image inpainting technique which is unable to preserve facial identity. Facial preservation is a must upon makeup application because in real condition, our face features will still be maintained even after makeup. Therefore, we benchmark the face identity preservation capability with the conventional image inpainting techniques, notably FcF\cite{fcf} and LaMa\cite{suvorov2022resolution}. To ensure the equitable comparisons, both baselines were trained on the BeautyFace dataset\cite{yan2023beautyrec}, adhering to identical training configurations as our proposed method.

\textbf{Evaluation Metrics} We employ two metrics for quantitative evaluations: \textbf{(i) Fréchet inception distance (FID)} \cite{heusel2017gans} the quality of images generated by machine learning models, particularly in the context of generative adversarial networks (GANs), where it measures the similarity between two sets of images, typically the set of real images and the set of images generated by the model. In our case, we utilize it to evaluate how well our model learns the makeup style from the dataset where it contains our desired style. It is evaluated by calculating and comparing the distribution of the generated makeup images $I_{\text{output}}$, with the distribution of the dataset containing our desired makeup style. \textbf{(ii) Additive Angular Margin Loss (ArcFace)} \cite{deng2019arcface}, normally used in scenarios where there is a need to verify or recognize faces with high accuracy, such as security systems, smartphone unlocking or identification processes. We utilize it to evaluate the face identity preservation capability of our model, determining the likelihood of our results matching the person’s identity in the makeup-free images. 

\textbf{Dataset.}
We trained \textit{Protégé} on BeautyFace dataset \cite{yan2023beautyrec}. This dataset features 2,719 high-resolution images that include various facial characteristics, expressions, poses, and lighting conditions. We used this dataset to represent a specific makeup style (i.e., basic makeup) where we expect our model to learn and generate makeups that follow the similar overall makeup style. We partitioned the dataset into a training set of 2,447 images and a test set of 272 images. 

\textbf{Implementation Details.}
Not all data augmentation techniques are suitable for our situation. In our case, we resize all the images to 256 x 256, adopting only a random flip as data augmentation. The random flip is applied simultaneously on both the input image $I_{\text{D}}$ and ROI Mask $M$. Specifically, when $I_{\text{D}}$ is flipped, $M$ must be flipped in the same direction. This is because $M$ is essential to be positioned the same as $I_{\text{D}}$ to indicate the correct region for makeup application. We employ Adam as optimizer with a learning rate of 0.00001 and a batch size of 2 on a Nvidia RTX 3070 GPU. Our model is implemented using PyTorch\cite{paszke2019pytorch}.

\subsection{Qualitative Results}
\textbf{Visual comparisons in makeup application between \textit{Protégé} and human-artist makeup.}
We visually compared \textit{Protégé} against the makeup applied by the human makeup artist. As shown in Fig. \ref{fig:qual_results}, our model demonstrates proficiency in generating makeup that is close to the general style of the makeups done by the human makeup artist (evaluated by FID). 

A notable observation in the results is \textit{Protégé}’s inclination towards pinkish hues, reflecting the overall style our model learns from the makeup set from human makeup artist $I_{\text{D}}$. This is evident in the lipstick shades in makeups done by \textit{Protégé}, where it prefers bright red to dark red, and favors pink lipstick color over nude lipstick color, showing a preference for vibrant shades. Nonetheless, \textit{Protégé} occasionally select more subdued nude tones due to the learned makeup distribution.

\begin{figure}[ht]
	\centering 
	\includegraphics[width=0.45\textwidth]{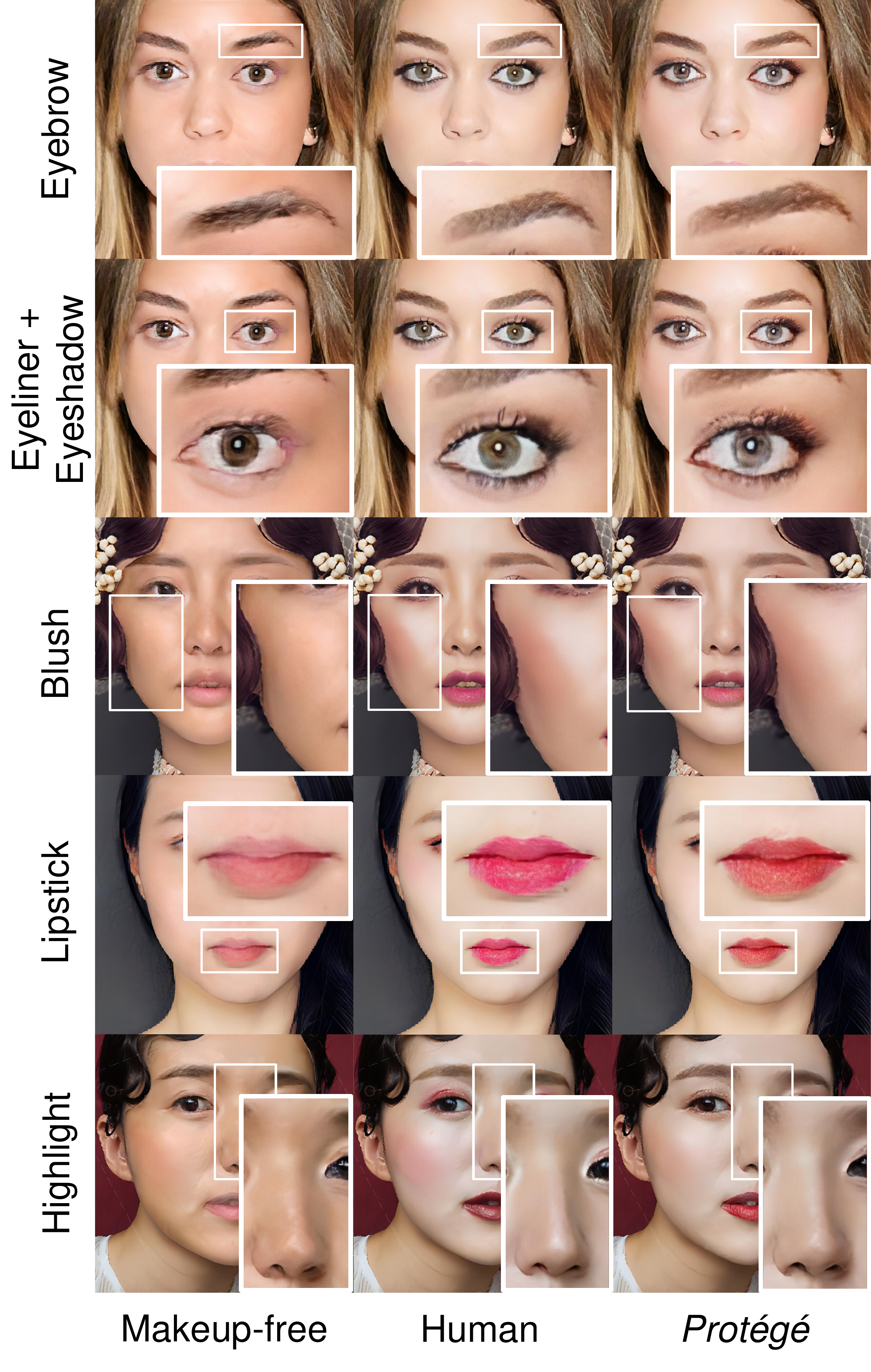}
	\caption{\textbf{Close-up views} of \textit{Protégé}'s makeups on \textit{makeup-free} faces such as eyebrows, eyeliner, eyeshadow, blush, and lipstick, highlighting its finesse and natural-looking outcomes.}
	\label{fig:details}
\end{figure}

\textit{Protégé}’s prowess in detailed makeup application is highlighted in Fig. \ref{fig:details}, showcasing \textit{Protégé}’s remarkable ability to capture the details from makeups done by human makeup artists $I_{\text{D}}$. 

\textbf{Visual Comparisons in Face Identity Preservation. Which of these faces belongs to the same person as the one shown in the makeup-free image?} 
\begin{figure}[ht]
	\centering 
	\includegraphics[width=0.5\textwidth]{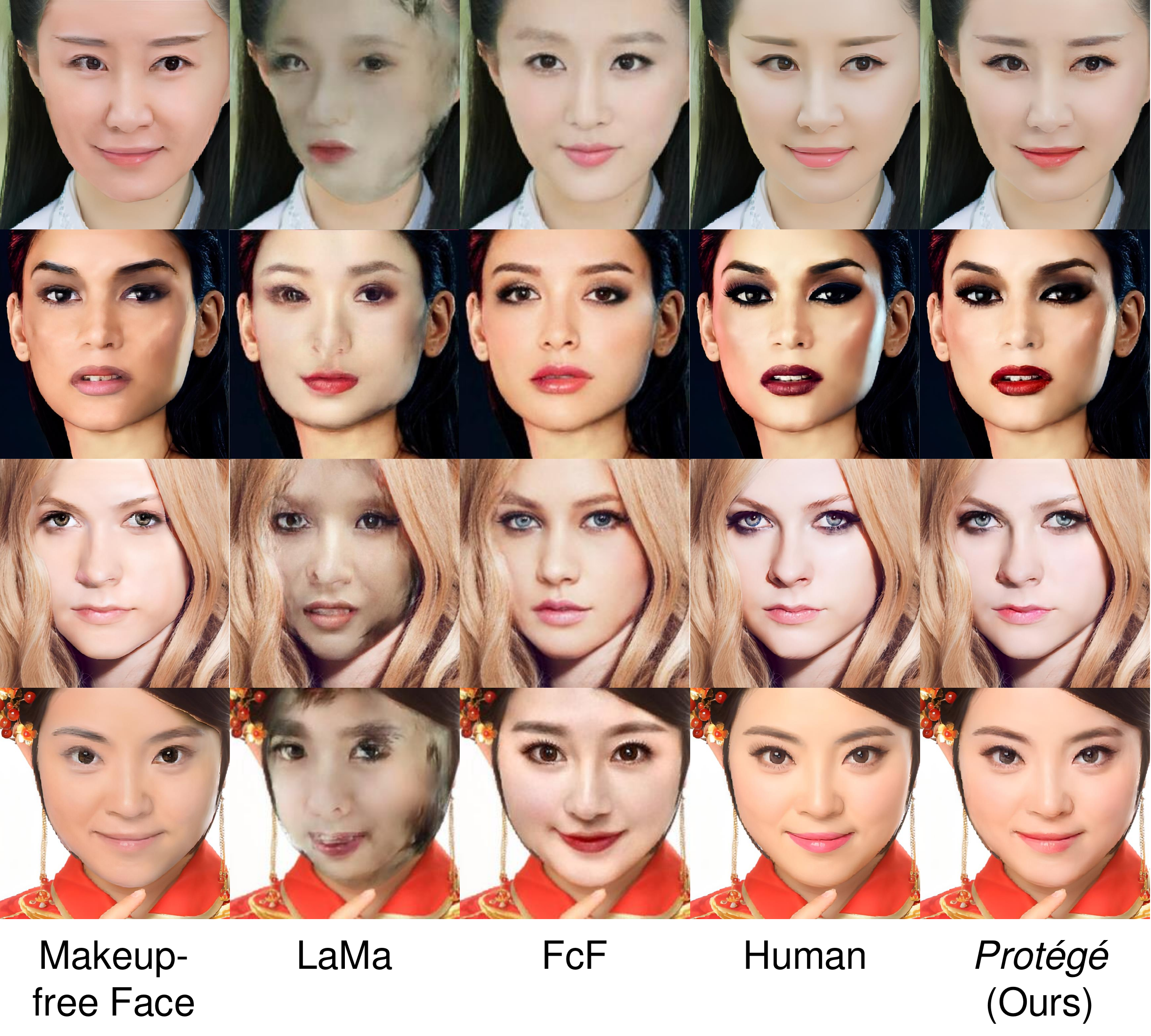}
	\caption{\textbf{Visual Comparisons in Face Identity Preservation:} Which of these faces belongs to the same person as the one shown in the \textit{makeup-free} image?}
	\label{fig:identity}
\end{figure}

We compared \textit{Protégé}’s resulting image after makeup with baseline models, which are FcF\cite{fcf} and LaMa\cite{suvorov2022resolution}. The results in Fig. \ref{fig:identity} demonstrate \textit{Protégé}’s superior capability in maintaining facial identity after makeup application, which is expected to be like how a human makeup artist would do-keep the face identity after makeup application. This capability stands \textit{Protégé} apart from the conventional image inpainting models. 

\subsection{Quantitative Results}

\begin{table}
    \centering
    \adjustbox{max width=\linewidth}{
        \begin{tabular}{l|cccc} 
        \hline
        Methods & FID$^1$ ($\downarrow$) & ArcFace$^{2}$ ($\uparrow$) \\
        \hline
        LaMa \cite{suvorov2022resolution} & 91.03 & 0.33 ($\pm$0.14) \\
        FcF \cite{fcf} & 40.00 & 0.59 ($\pm$0.14)\\
        \textbf{\textit{Protégé}} (Ours)    & \bf 27.26 & \bf 0.89 ($\pm$0.07)  \\
        \hline
        \end{tabular}
    }
            \caption{\textbf{Quantitative Comparisons of \textit{Protégé}:} Evaluating Makeup Style Learning$^1$ and Face Identity Preservation$^2$ between $I_{\text{output}}$ and $I_{\text{D}}$ which correspond to makeup images generated by \textit{Protégé} v.s. makeup image by human makeup artist. 
    }
    \label{tab:quan_score}
\end{table}

Apart from qualitative evaluation, we further evaluate our model in the makeup style learning capability and facial identity preservation.

\textbf{How well did our model \textit{Protégé} learn makeup style from $I_{\text{D}}$?} 
As shown in Tab.~\ref{tab:quan_score}, our \textit{Protégé} outperforms LaMa and FcF in generating style-aligned makeups as supported by the lowest FID score (\textit{i.e.}, 27.26).
 
\textbf{How well did our model \textit{Protégé} preserve face identity during makeup application?} 
The finding in ArcFace score as shown in Tab.~\ref{tab:quan_score} shows that \textit{Protégé} effectively maintains face identity with the highest ArcFace score (i.e., 0.89), despite the slight variations due to different makeup applied. 

The results demonstrate that our model effectively learns the makeup style and proficiently generates makeups, as assessed in qualitative evaluations, while maintaining the integrity of the facial identity.

\subsection{User Study}
\begin{table}
    \centering
    \begin{tabular}{l|cc} 
    \hline
    Methods & Prefer Makeup by? & Identical to \( I_{\text{D}}? \) \\
    \hline
    LaMa \cite{suvorov2022resolution} & - & 0.00\% \\
    FcF \cite{fcf} & - & 5.00\%\\
    Human Artists & 25.00\% & 100.00\%\\
    \textbf{\textit{Protégé}} (Ours) & 75.00\% & 100.00\%\\
    \hline
    \end{tabular}
    \caption{\textbf{User Study:} Comparing User Preferences for Makeups between \textit{Protégé} and Humans and Assessing \textit{Protégé}'s Facial Identity Preservation Capability.}
    \label{tab:userstudy}
\end{table}

We evaluated subjective preferences and perceptions from humans for \textit{Protégé}’s generated makeups. The participants were given a set of 20 questions, where each question is presented with a face alongside its two makeup versions, one is makeup face done by \textit{Protégé}, while another is makeup face done by the human makeup artist. However, the information of whether the makeup is done by \textit{Protégé} or human makeup artist is not exposed to the participants. In the situation of believing both options were from the makeup application models, they were not aware which one was makeup done by human makeup artist. 

From the user study shown in Tab.~\ref{tab:userstudy}, we are surprised that many favored makeup applications done by the \textit{Protégé} with a percentage of 75.00\%. This is under the impression that both options were generated by \textit{Protégé}, specifically, the participants did not expect one of the options is makeup done by human makeup artist. We also conducted interview to these participants to learn the reason of their choices. We are informed that the reason is mostly because of the makeups generated by \textit{Protégé} is more natural and has a closer resemblance to common human’s makeup styles. This preference suggests that \textit{Protégé} not only effectively captures and learns from the makeup dataset done by human makeup artist, but also surpassed our expectations. 

Moreover, we conducted survey related to the capability of \textit{Protégé} to preserve face identity upon makeup application. In the survey, makeup-free faces were shown alongside four options: (i) Makeup image by \textit{Protégé}, (ii) Makeup image by human makeup artist, (iii) Makeup image by FcF, and (iv) Makeup image by LaMa. Participants had to identity which makeup image is the person similar to the makeup-free face presented. The results show that many participants have linked our makeup image with the original face, showing our model’s success in preserving face identity. 

\section{Conclusion}

This study successfully overcomes the limitations of conventional makeup application methods such as manual makeup application, makeup transfers, and rule-based makeup recommendation systems using expert system, which often lack the capability to create intuitive and innovative makeup, causing limitation in fulfilling the increasing demands of users on the digital platform. Traditional approaches not only require extensive user interaction and makeup knowledge, but they are also constrained by static rules and pre-existing makeup styles, which hinder the creation of innovative makeup looks. Our innovative model, \textit{Protégé}, introduces a transformative makeup application model. By leveraging our advanced makeup inpainting algorithm, \textit{Protégé} simplifies the learning and application processes, enabling intuitive and bespoke makeup creations that maintain the user’s facial identity. Comprehensive testing in the BeautyFace dataset has demonstrated \textit{Protégé}’s ability to deliver personalized, diverse, and high-quality makeup efficiently on subject’s face, redefining user experience in digital basic makeup application.

{\small
\bibliographystyle{IEEEbib}
\bibliography{IEEEbib_ref}
}






\end{document}